\documentclass[10pt,twocolumn,letterpaper]{article}
\usepackage{geometry}
\usepackage{cvpr}
\usepackage{times}
\usepackage{epsfig}
\usepackage{graphicx}
\usepackage{amsmath}
\usepackage{amssymb}

\usepackage{amsfonts}
\usepackage{subfigure}
\usepackage{amsmath}
\usepackage{amssymb}
\usepackage{array}
\usepackage{booktabs}
\usepackage{multirow}
\usepackage{color}
\usepackage{algorithm}
\usepackage{algorithmicx}
\usepackage{algpseudocode}
\usepackage{amsmath}
\usepackage{verbatim}
\usepackage{psfrag,epsfig,epstopdf}
\usepackage[normalem]{ulem}
\usepackage[noblocks]{authblk}

\newcommand{\jk}[1]{{\color{black}#1}}
\newcommand{\jkk}[1]{{\color{black}#1}}
\newcommand{\jkkk}[1]{{\color{black}#1}}
\newcommand{\jingk}[1]{{\color{black}#1}}
\newcommand{\jingke}[1]{{\color{black}#1}}
\newcommand{\ws}[1]{{\color{black}#1}}
\newcommand{\wss}[1]{{\color{black}#1}}
\newcommand{\meng}[1]{{\color{black}#1}}
\newcommand{\wsh}[1]{{\color{black}#1}}
\newcommand{\wush}[1]{{\color{black}#1}}
\newcommand{\wusheng}[1]{{\color{black}#1}}
\newcommand{\jingkee}[1]{{\color{black}#1}}
\newcommand{\jingkeee}[1]{{\color{black}#1}}
\newcommand{\mjk}[1]{{\color{black}#1}}
\newcommand{\final}[1]{{\color{black}#1}}
\newcommand{\erhao}{\fontsize{21pt}{\baselineskip}\selectfont}

\usepackage[breaklinks=true,bookmarks=false]{hyperref}

\cvprfinalcopy 

\def\cvprPaperID{964} 

\ifcvprfinal\fi
\begin{document}

{\onecolumn

\noindent \textbf{\erhao{Weakly Supervised Person Re-Identification}}

\vspace{2cm}

\noindent {\LARGE{Jingke Meng, Sheng Wu, Wei-Shi Zheng}}

%

\vspace{1cm}

\noindent For reference of this work, please cite:

\vspace{1cm}
\noindent Jingke Meng, Sheng Wu, Wei-Shi Zheng.
``Weakly Supervised Person Re-Identification.''
In \emph{Proceedings of the IEEE International Conference on Computer Vision and Pattern Recognition.} 2019.

\vspace{1cm}

\noindent Bib:\\
\noindent
@inproceedings\{meng2019weakly,\\
\ \ \   title=\{Weakly Supervised Person Re-Identification\},\\
\ \ \  author=\{Meng, Jingke and Wu, Sheng and Zheng, Wei-Shi\},\\
\ \ \  booktitle=\{Proceedings of the IEEE International Conference on Computer Vision and Pattern Recognition\},\\
\ \ \  year=\{2019\}\\
\}
}

%
\restoregeometry
\title{Weakly Supervised Person Re-Identification}

\author[ ]{\vspace{-0.5cm}Jingke Meng$^{1,3}$}
\author[ ]{Sheng Wu$^{1}$\vspace{-0.3cm}}
\author[ ]{Wei-Shi Zheng$^{1,2}$\thanks{Corresponding author}}

\affil[ ]{$^{1}$\small School of Data and Computer Science, Sun Yat-sen University, China}
\affil[ ]{$^{2}$\small Key Laboratory of Machine Intelligence and Advanced Computing, Ministry of Education, China}
\affil[ ]{$^{3}$\small Accuvision Technology Co. Ltd, China}
\affil[ ]{\small mengjke@mail2.sysu.edu.cn, wush43@mail2.sysu.edu.cn, wszheng@ieee.org \vspace{-0.3cm}}

\maketitle
\thispagestyle{empty}

\begin{abstract}
In the conventional person re-id setting, it is assumed that the labeled images are the person images within the bounding box for each individual; this labeling across multiple nonoverlapping camera views from raw video surveillance is costly and time-consuming. To overcome this difficulty, we consider weakly supervised person re-id modeling. The weak setting refers to matching a \jingke{target person} with an untrimmed gallery video where \wusheng{we} only know that the identity appears in the video \wusheng{without \mjk{the} requirement of annotating the identity in any frame of the video during the training procedure.}
Hence, for a video, there could be multiple \jingkee{video-level} labels. We cast this weakly supervised person re-id challenge into a multi-instance multi-label learning (MIML) problem.
In particular, we develop a \textit{Cross-View MIML (CV-MIML)} method that is able to explore potential intraclass \jingke{person images} \mjk{from all the camera views by incorporating the intra-bag alignment and the cross-view bag alignment.}
Finally, the CV-MIML method is embedded into an existing deep neural network for developing the \textit{Deep Cross-View MIML (Deep CV-MIML)} model.
We have performed extensive experiments to show the feasibility of the proposed weakly supervised \mjk{setting} and verify the effectiveness of our method compared to related methods on four weakly labeled datasets.
\end{abstract}

\section{Introduction}

Given an image from a set of probe images, the objective of person re-identification (re-id) is to identify the same person \jkkk{across a set of gallery images from nonoverlapping camera views}. The changes in illumination, camera viewpoint, background and occlusions lead to considerable visual ambiguity and appearance variation and make person re-id a challenging problem. Several representative methods \cite{Yang2018video,wu2018exploit,zhu2018video,li2018diversity} have been developed to solve this problem.
\begin{figure}[t]
  \centering
  \subfigure[Conventional fully supervised setting]{
    \label{fig:introduction_a} 
    \includegraphics[width=\columnwidth]{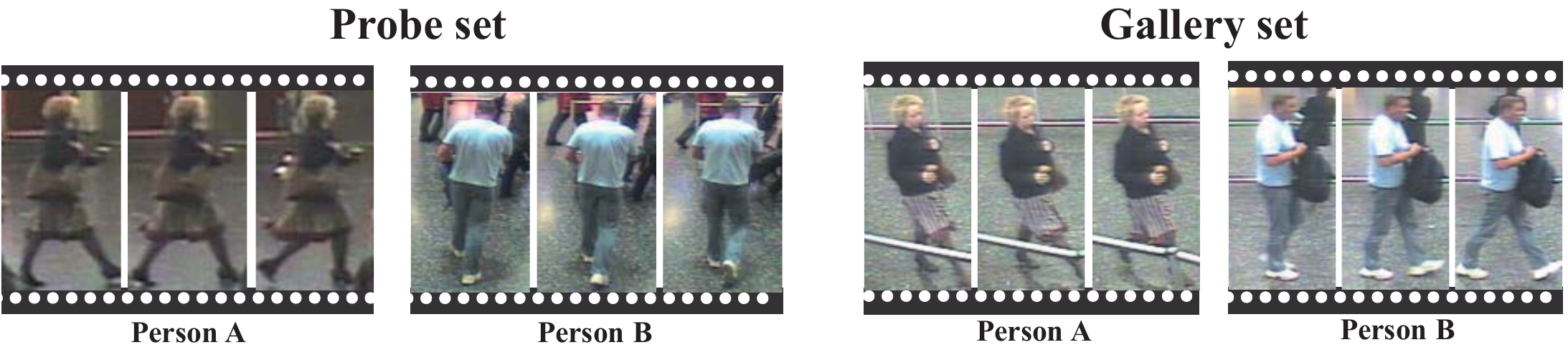}}
  \hspace{1in}
  \subfigure[Proposed weakly supervised setting]{
    \label{fig:introduction_aa} 
    \includegraphics[width=\columnwidth]{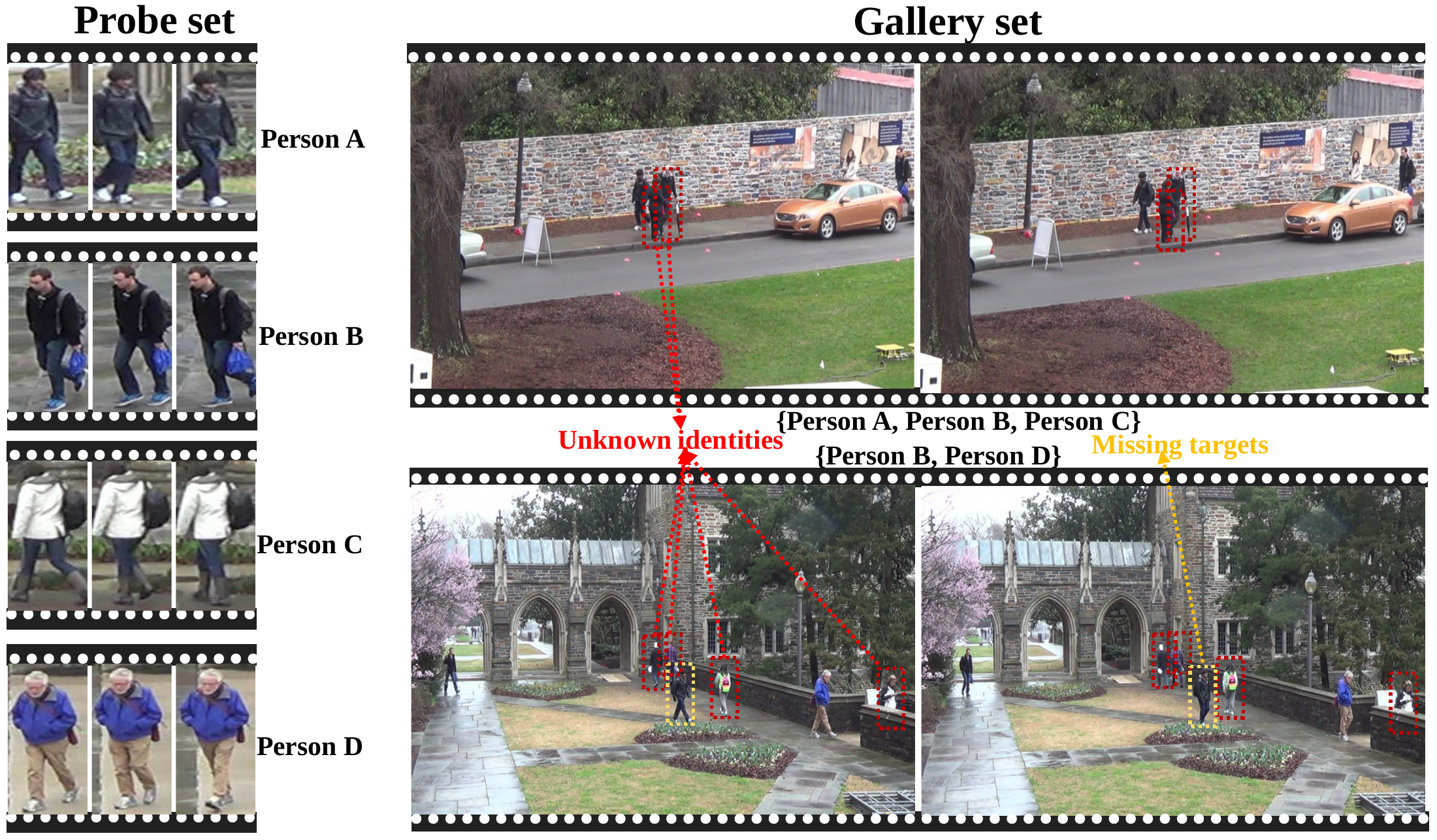}}
  \caption{Comparison of two settings. (a) Conventional fully supervised setting: image sequences in the probe and gallery set are manually trimmed and labeled from video surveillance in a frame-by-frame manner.
   (b) Proposed weakly supervised setting: \ws{the untrimmed videos in the gallery set are tagged by multiple video-level labels, while the specific label of each individual is absent from the labeling process.}}
  \label{fig:subfig} 
  \vspace{-0.2cm}
\end{figure}

While numerous methods have been developed for \jingke{fully supervised} person re-id, conventionally, it is assumed that for model training, 1) the images in the \mjk{probe set} and \mjk{gallery set} are manually trimmed and labeled from raw video surveillance (probably with the assistance of detection) frame-by-frame (as shown in Figure~\ref{fig:introduction_a}), and 2) all training samples are of the target to be matched, and no outliers exist.
Although such precise annotations could eliminate the difficulty of learning robust person re-id models, they require strong supervision, which makes the entire learning process difficult \wsh{to adapt} to large-scale person re-id in a more practical and challenging scenario.

Instead of relying on costly labeling/annotations, we wish to investigate the person re-id modeling in a weakly supervised setting.
This setting assumes that
annotators only need to take a \wsh{rough} glance at the raw videos to determine \mjk{which identities} appear in such videos, and they do not need to annotate the \mjk{identity} in any frame of the video. That is, only the video-level label indicating the presence of the \mjk{identity} is given, while the ground-truth regarding in which frame and which bounding box in a frame the \mjk{identity} is present is not provided.
In such a setting, the labeling cost of person re-id can be greatly reduced \jingke{compared to the conventional fully supervised setting}.
We call this setting \textbf{\textit{weakly supervised person re-id}}.


More specifically, as shown in Figure~\ref{fig:introduction_aa}, the first \wsh{row of} a video clip in the gallery set \wsh{is} annotated with a set of video-level labels \{Person A, Person B, Person C\} indicating that Person A, Person B and Person C have appeared in this video clip, but there is no \wsh{additional prior knowledge} that \wsh{precisely indicates} which individual is Person A, Person B or Person C. Hence, these labels are weak. Note that it is possible that some labels for a video are missing because the annotators fail to recognize (\eg, pedestrians framed by yellow dotted lines in Figure~\ref{fig:introduction_aa}). It is also practically possible that unknown identities appear in the untrimmed video clips (\eg, pedestrians framed by red dotted lines in Figure~\ref{fig:introduction_aa}). Overall,
the videos in the gallery set are untrimmed and tagged with the multiple video-level weak labels in this weakly supervised setting. Based on this setting, we aim to
find in the gallery the raw videos where the target person appears, given a probe set of images from nonoverlapping camera views.


To solve the problem of weakly supervised person re-id, we consider every video clip in the gallery set as a bag; each bag contains multiple instances of the person images detected in each raw video clip \jingkee{and associates with multiple bag-level labels}.
For the probe set, it contains the target individuals to be searched for in the gallery; thus, each input is a set of manually trimmed images of the target person.
For convenience, we also regard the probe input as a bag.
We consider the whole weakly supervised person re-id problem as a multi-instance multi-label learning (MIML) problem and develop a \emph{Cross-View MIML (CV-MIML) \mjk{method}}. Compared to existing MIML algorithms~\cite{briggs2013instance,briggs2013context,huang2014fast,hu2012towards,pham2015multi,zhu2017discover,feng2017deep}, \mjk{our} CV-MIML is \mjk{able} to exploit similar instances \jingke{within a bag} for intra-bag alignment and mine potential matched instances between bags that are captured across camera views through embedding distribution prototype into MIML, which is called the cross-view bag alignment in our modeling.
Finally, we embed this CV-MIML method into a deep neural network to form an end-to-end
{deep cross-view multi-label multi-instance learning (Deep CV-MIML) \mjk{model}.


To the best of our knowledge, this paper is the first to propose and study the weakly supervised problem in person re-id. We have performed comprehensive experiments on four datasets with one genuine dataset and three simulated datasets. Since existing person re-id methods do not suit the weakly supervised setting, we  compare the proposed method to other state-of-the-art MIML methods and several state-of-the-art one-shot, unsupervised and sully supervised person re-id methods.
The results  demonstrate \jingke{the feasibility of the weakly supervised person re-id method} and show that the proposed Deep CV-MIML \mjk{model} is a superior approach to solving the problem.

\section{Related Work}

\subsection{Person Re-identification}

Most studies of person re-id are supervised \cite{zhou2018deep,wu2018deep,cheng2018deep,Yang2018video,wu2018exploit,zhu2018video,li2018diversity,xu2018attention,song2018mask,chen2018person} and require annotating each person in the video precisely (e.g., indicating the frame and the position in the frame within the video).
\jingke{
It is impractical to extend to the above person re-id methods in a more practical and challenging scenario due to the expensive cost of the labeling process. So we propose the weakly supervised setting for person re-id which only requires video-level weak labels.
}

Recently, several \mjk{unsupervised learning methods} have been \mjk{developed} to learn person re-id models \cite{yu2017cross,fan2017unsupervised,ma2017person,liu2017stepwise,ye2017dynamic,li2018unsupervised,chen2018deep}. The general idea of these methods is to \mjk{explore} unlabeled data progressively by alternately assigning pseudo-labels to unlabeled data and updating the model according to these pseudo-labeled data.
The unsupervised learning process can be easily adapted to large-scale person re-id since the unlabeled data can be accessed without manual operations.
However,
the performance of these unsupervised methods is limited because
the visual ambiguity and appearance variations are not easy to address due to the lack of clear supervised information.

\jingke{In the weakly supervised setting, the gallery set is composed of the raw videos, which is closely related to the person search \cite{xiao2016end} that aims to search for the target person from the whole raw images.}
However, \mjk{in the} setting \mjk{of} the person search, the manually annotated bounding boxes for the gallery set are required to train the model in a fully supervised manner, which is much more supervised than our weakly supervised setting.



\subsection{Multi-Instance Multi-Label Learning}\label{miml}
\jingke{
In general, an object of interest has its inherent structure and it can be represented as a bag of instances with multiple labels associated on the bag level. Multi-Instance Multi-Label learning (MIML) \cite{zhou2012multi} provides a framework for handling this kind of problems.}}
\wsh{Due to the limitation of the current person re-id methods in the weakly supervised setting, we adopt the MIML formulation to solve our weakly supervised re-id problem.
During the past few years, many related algorithms have been {investigated and developed for} MIML problems \cite{briggs2013instance,briggs2013context,huang2014fast}.
The MIML formulation has also been applied in many practical vision domains, such as image annotation \cite{xu2011ensemble,nguyen2013multi} and classification tasks~\cite{yakhnenko2011multi,chen2013multi,yang2017miml,zha2008joint}.}



While it is possible to apply existing MIML to our problem, there still exist several intractable issues that may
not be readily resolved
because of the following: 1) the existing MIML methods ignore mining the intra-bag variation between similar instances \jingke{belonging to the same person};
2) previous approaches are based on the idea that highly relevant labels mean sharing \mjk{common instances} among the corresponding classes, but the class labels are independent from each other in person re-id; and
3) \wss{most MIML methods are not able to mine potential matched
instances between bags effectively when applied to person re-id for cross-view matching.}
\jingke{
The proposed Deep Cross-View MIML \mjk{model} for the person re-id can overcome the above limitations by exploiting similar instances \mjk{within a bag} for intra-bag alignment and mining potential matched instances across camera views simultaneously.
}


\section{The Proposed Approach}

In this section, we formally introduce the weakly supervised person re-id setting
and then introduce the Deep CV-MIML model for addressing this problem.

\subsection{Problem Statement and Notation}
In the weakly supervised person re-id setting, our goal is to find the videos that the target person appears in, given a probe set of images from nonoverlapping camera views.
Suppose that we have $C$ known identities from $V$ camera views and that every known identity appears in at least two camera views. Since some unknown identities \jingke{(\eg, pedestrians framed by red dotted lines in Figure \ref{fig:introduction_aa})} would appear in the untrimmed videos, these unknown identities can be affiliated to a novel class; we define an extra 0-class to
represent it. For simplicity, we denote the overall number of classes by $\tilde{C}=C+1$.

In our learning, given $N_{\mathcal{X}}$ videos, the training set $\mathcal{X}$ consists of two distinct parts: the probe set $\mathcal{X}_p$ and the gallery set $\mathcal{X}_g$.
The videos in the gallery set are untrimmed and tagged with the multiple video-level weak labels that indicate the presence of individuals as shown in Figure~\ref{fig:introduction_aa}; the person images within a raw video in the gallery set are automatically detected
in advance.
Note that even though the person images are detected during this stage, the \wsh{specific} label of each individual is still \wsh{unknown}.


We consider every raw video as a bag; each bag contains multiple instances of the person images detected in each video. For the probe \wsh{set, each query is composed of a set of} detected images of the same person. For convenience, we also regard \wsh{each query in the probe set} as a bag.
More
specifically, the training set can be denoted by $\mathcal{X} = \{\mathcal{X}_p, \mathcal{X}_g \}$, where the probe set is $\mathcal{X}_p=\{(X_b,\mathbf{y}_b,v_b)\}_{b=1}^{N_p}$ and the gallery set is $\mathcal{X}_g=\{(X_b,\mathbf{y}_b,v_b)\}_{b=1}^{N_g}$,
$N_{\mathcal{X}} = N_{p} + N_{g}$.
\wsh{For the bags (videos) in the probe set, each} bag $X_b$ \wsh{containing the same person images} is labeled
by $\mathbf{y}_b$ under the $v_b$-th camera view, where $v_b \in \{1,2,...,V \}$, and $\mathbf{y}_b=[y_{b}^0,y_{b}^1,...,y_{b}^C] \in \{0,1\}^{\tilde{C}} $ is a label vector containing ${\tilde{C}}$ class labels, in which
$y_{b}^c=1$ if the $c$-th label is tagged for $X_b$, and $y_{b}^c=0$ otherwise.
In contrast to the conventional person re-id, \jingk{for the \wsh{bags (videos)} in the gallery set}, \wsh{$y_{b}^c=1$ denotes that the $c$-th identity appears in this \wsh{bag (video)}, while} $y_{b}^c=0$ denotes uncertainty of whether the $c$-th identity has appeared in this video. 
Moreover, the bag $X_b$ consists of $n_b$
instances $\mathbf{x}_{b,1}$, \wsh{$\mathbf{x}_{b,2}$, ..., $\mathbf{x}_{b,{i}}, ...,\mathbf{x}_{b,n_b}$}, where $\mathbf{x}_{b,{i}} = f_e(\mathbf{I}_{b,i};\theta)\in \mathbb{R}^d$
is the feature vector extracted from the corresponding person image $\mathbf{I}_{b,i}$, and $f_e(\cdot;\theta)$ is a learnable feature extractor.

\subsection{Cross-View MIML for Person Re-id}
We cast the weakly supervised person re-id as the problem of multi-instance multi-label learning (MIML) and present the cross-view multi-instance multi-label (CV-MIML) learning method to solve this problem.

\subsubsection{Weakly Supervised Person Re-id by MIML}
For the task of weakly supervised classification, we formulate a MIML classifier for our weakly supervised person re-id.
With this classifier $f_c(\cdot;W)$, the high-dimensional input $\mathbf{x}_{b,i} \in \mathbb{R}^d $ can be transformed into a $\tilde{C}$-dimensional vector $\tilde{\mathbf{y}}_{b,i} = f_c(\mathbf{x}_{b,i};W) \in \mathbb{R}^{\tilde{C}}$ that can be interpreted as a label distribution, embedding the similarities among all classes.

For the probe set $\mathcal{X}_p$, all instances \jingke{$\{\mathbf{x}_{b,i}\}_{i=1}^{n_b}$}  in each bag $X_b$ are tagged with the same label $\mathbf{y}_b$. For the gallery set $\mathcal{X}_g$, all instances $\{\mathbf{x}_{b,i}\}_{i=1}^{n_b}$ in each bag $X_b$  share the same weak video-level label $\mathbf{y}_b$. The softmax classifier cannot be directly applied to \jingkee{the instances in the gallery set} because the specific label of each instance is absent.
Therefore,
the instances from the probe set $\mathcal{X}_p$ and gallery set $\mathcal{X}_g$ are separately processed by the following two \jingkee{procedures} to learn a classification model.

\wsh{On the one hand, we expect the estimated label distribution to eventually approximate the true one;} thus, the classification loss for these instances in the probe set can be written as follows:
\begin{equation}\label{eq:class_loss1}
\begin{aligned}
\mathcal{L}_{p} =& \frac{1}{N_{p}} \sum \limits_{X_{b} \in \mathcal{X}_p} \sum \limits_{i\in \{1,\cdots,n_b\}} \sum \limits_{c \in \{0,\cdots,C\}} (-{y}_{b}^c \log \tilde{y}_{b,i}^c ),
\end{aligned}
\end{equation}
where $y_{b}^c$ denotes the ground truth video-level labels of bag $X_b$ at \wsh{the} $c$-th entry, $\tilde{y}_{b,i}^c$ is the $c$-th estimated probability of the $i$-th instance in bag $X_b$, and $N_{p}$ indicates the overall number of instances involved in the loss calculation.

\begin{figure}[t]
  \centering
  {
    \includegraphics[width=\columnwidth]{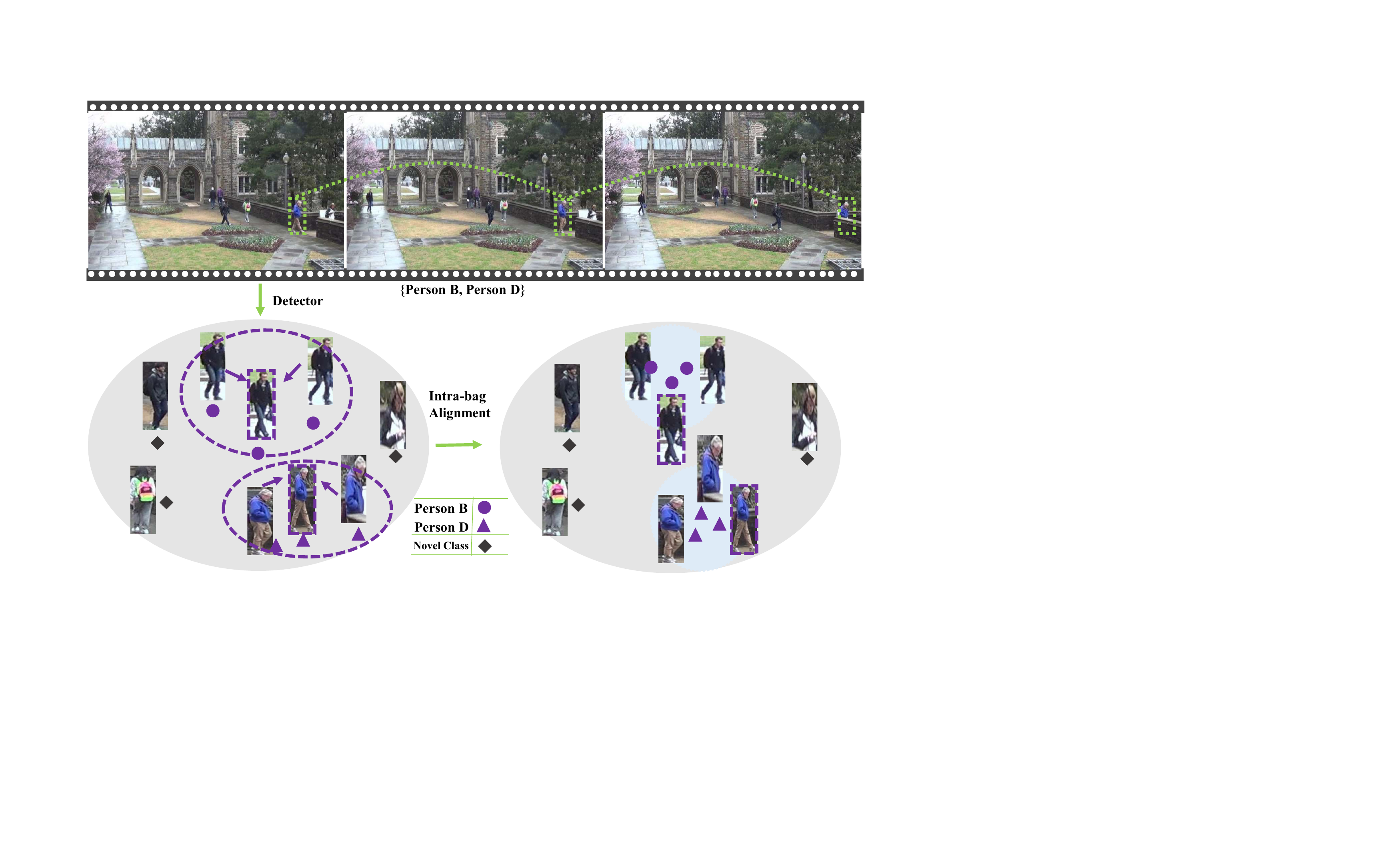}}
  \caption{Illustration of the intra-bag alignment. \jingkee{The instances in the rectangular with dotted purple lines are the seed instances corresponding to the Person B and Person D, respectively. Then two groups
(\eg, framed by the purple oval dotted lines) are formed around these two seed instances. In the intra-bag alignment process, the label distributions of instances belonging to the same group are aligned such that these instances can be compact between each other in the learned feature space.}
%
}
  \label{fig:intra} 
  \vspace{-0.2cm}
\end{figure}

\wsh{On the other hand, we further expect that our classifier can fully exploit the weak labels to learn a more robust re-id model.
\jingkeee{More specifically, for any tagged label $c$ in bag $X_b$, we select \wsh{an} instance with the largest prior probability w.r.t the $c$-th class as the \textit{seed instance} $\mathbf{x}_{b,q_c}$, where the index $q_c$ can be defined by
\begin{equation}\label{eq:seed}
q_c = \text{argmax}_{i\in\{1,2,\cdots,n_b\}} \{\tilde{y}_{b,i}^c\}.
\end{equation}
Then we}
\jingke{force the estimated label of the seed instance approximate to the corresponding tagged video-level label.}
Accordingly, we define the classification loss for the gallery set as follows:
\begin{equation}\label{eq:class_loss2}
\begin{aligned}
\mathcal{L}_{g} = \frac{1}{N_{g}} \sum \limits_{X_{b} \in \mathcal{X}_g} \sum \limits_{c \in \{0,\cdots,C\}} (-{y}_{b}^c \log \max \limits \{\tilde{y}_{b,1}^c,\tilde{y}_{b,2}^c,...,\tilde{y}_{b,n_b}^c\} ),
\end{aligned}
\end{equation}
where the operation $\max \limits \{\tilde{y}_{b,1}^c,\tilde{y}_{b,2}^c,...,\tilde{y}_{b,n_b}^c\}$ is used to select the largest prior probability of the seed instance \jingkeee{$\mathbf{x}_{b,q_c}$.}
In such a case, the classification model can be leveraged to infer the prior probability of each instance in the bag.

Combining the \mjk{two} classification \mjk{losses} for the probe set (Eq.(\ref{eq:class_loss1})) and the gallery set (Eq.(\ref{eq:class_loss2})), we obtain the \wsh{following} MIML classification loss:
\begin{equation}\label{eq:class_loss}
\begin{aligned}
\mathcal{L}_{C} = \mathcal{L}_{p} + \mathcal{L}_{g}\mjk{.}
\end{aligned}
\end{equation}

\subsubsection{Intra-bag Alignment}
\jk{
Since individuals often appear in a video across several consecutive frames (\eg, green dotted lines in Fig. \ref{fig:intra}), there will be a set of instances, probably of the same person, in a bag in the weakly labeled gallery set. \jingkee{These instances are expected to be merged} into a group \wsh{such} that the instances belonging to the same group should be close to each other in the learned feature space.
However, the MIML classifier cannot achieve this \wsh{agglomeration} and the classifier only processes the instance with the largest prior probability w.r.t the corresponding classes, which we call the \emph{seed instance}.

\wsh{To this end,
we expect that \jingkee{the set of} instances probably of the same person can be gathered around the seed instance} $\mathbf{x}_{b,q_c}$
that has the largest prior probability with respect to the $c$-th class \jingkee{in the bag $X_b$}.
Then, we form a group that contains the instances gathered around the seed instance $\mathbf{x}_{b,q_c}$ by
$\mathcal{G}_{b,c} = \{p|\mathbf{x}_{b,p} \in \mathcal{N}_{q_c} \text{ and } \tilde{{y}}_{b,p}^c \geq \gamma  \tilde{{y}}_{b,q_c}^c\}$. In this group, the \jingkee{selected} instances should be among the K-nearest neighbors $\mathcal{N}_{q_c}$ in the feature space \wsh{around} the seed instance $\mathbf{x}_{b,q_{c}}$. Additionally, the prior probability corresponding to the $c$-th class of these instances should be \mjk{no} less than $\gamma \tilde{y}_{b,q_c}^c$, where $\tilde{y}_{b,q_c}^c$ is the prior probability of the corresponding seed instance.
Here, $\gamma\in(0,1)$ is a relaxation parameter.}
Then, the intra-bag alignment loss can be defined as follows:
\meng{\begin{equation}\label{eq:ica}
\mathcal{L}_{IA}= \frac{1}{N_{IA}} \sum \limits_{X_{b} \in \mathcal{X}_g} \sum \limits_{c \in \{0,\cdots,C\}}  \sum \limits_{p \in \mathcal{G}_{b,c}}y_b^c D_{KL}(\tilde{\mathbf{y}}_{b,p}\|\tilde{\mathbf{y}}_{b,q_c}),
\vspace{-0.1cm}
\end{equation}
\begin{equation}\label{eq:kl}
D_{KL}(\tilde{\mathbf{y}}_{b,p}\|\tilde{\mathbf{y}}_{b,q_c}) = \sum \limits_{c \in \{0,\cdots,C\}} \tilde{y}_{b,p}^c (\log \tilde{y}_{b,p}^c - \log \tilde{y}_{b,q_c}^c).
\end{equation}}
\jingkee{{The intra-bag alignment loss in Eq.(\ref{eq:ica}) is designated to evaluate the discrepancy of the label distribution between the instances within the group $\mathcal{G}_{b,c}$ and the corresponding seed instance $\mathbf{x}_{b,q_c}$}.
The discrepancy between two label distributions is defined by the Kullback-Leibler divergence depicted in Eq.(\ref{eq:kl}).}
As \wsh{illustrated} in Figure \ref{fig:intra}, by minimizing the intra-bag alignment loss, the features of the same \meng{group} can become closer to each other due to the alignment between potential instances of the same class in a bag.

\subsubsection{Cross-view Bag Alignment}
The intra-bag alignment term mainly considers the person images that appear in the same bag.
We further expect to mine potential matched images of the same person between bags not only from the same
camera view but also from non-overlapping camera views.
In the meantime, all instances belonging to the same person
should form a compact cluster in the learned feature space.
\jingkeee{For this purpose, we introduce a distribution prototype for each class, and then all the potential matched images of the same person from all the camera views are expected to be aligned to the corresponding distribution prototype. Formally, the distribution prototype of the $c$-th class at the current epoch $t$ is denoted by $\hat{\mathbf{p}}_{c}^t$ that can be calculated by
\begin{equation}\label{eq:c_pct}
{\mathbf{p}}_{c}^t = \frac{1}{|\mathcal{V}_c|} \sum \limits_{v \in \mathcal{V}_c} (\frac{1}{|\mathcal{I}_{c,v}|}  \sum \limits_{i \in \mathcal{I}_{c,v}} \tilde{\mathbf{y}}_i),
\end{equation}
\begin{equation}\label{eq:up_pct}
\hat{\mathbf{p}}_{c}^t = \alpha \hat{\mathbf{p}}_{c}^{t-1} + (1-\alpha) {\mathbf{p}}_{c}^t,
\end{equation}
where
$\mathcal{V}_c$ is the collection of all the camera views, $\mathcal{I}_{c,v}$ is the
set of instance index\wsh{es} that belong to the $c$-th class under \ws{the} $v$-th camera view,
and $\alpha$ is a smoothing hyperparameter that controls the weight of the historical distribution prototype \meng{$\hat{\mathbf{p}}_{c}^{t-1}$ at the previous epoch $t-1$ when updating the distribution prototype at current epoch $t$}.
}

After that, we alternate between the following \wsh{two} steps in the training stage:
1) \wsh{calculate} the distribution prototype at current epoch $t$ for each class based on Eq. \wsh{(\ref{eq:c_pct})} and Eq. \wsh{(\ref{eq:up_pct})}; 2) \wsh{align} the label distributions of instances belonging to the same person from all the camera views to the corresponding  distribution prototype.
Specifically, the \textit{Cross-view Bag Alignment} is defined by
\meng{
\begin{equation}\label{eq:cca}
\begin{aligned}
\mathcal{L}_{CA}= \frac{1}{N_{CA}} \sum \limits_{X_b \in (\mathcal{X}_p \bigcup \mathcal{X}_g) } \sum \limits_{c \in \{0,\cdots,C\}} \sum \limits_{i \in \mathcal{I}_{c}}  y_b^c D_{KL}(\tilde{\mathbf{y}}_{b,i}\|\hat{\mathbf{p}}_{c}^t ),
\end{aligned}
\end{equation}
}
where $\mathcal{I}_{c}$ is the collection of instance index\wsh{es} from all the camera views for the $c$-th class\wsh{,}
and \jk{$\hat{\mathbf{p}}_{c}^t$ is the distribution prototype of the $c$-th class at the current epoch $t$.
\wss{As \wsh{illustrated} in Figure \ref{fig:cross}, $\mathcal{L}_{CA}$ is minimized to make potential instances of the same person from different bags captured from different camera views \wsh{to gather} together.
}}


\begin{figure}[t]
\setlength{\belowcaptionskip}{-0.5cm}
  \centering
  {
    \label{fig:introduction_b} 
    \includegraphics[width=0.5\textwidth]{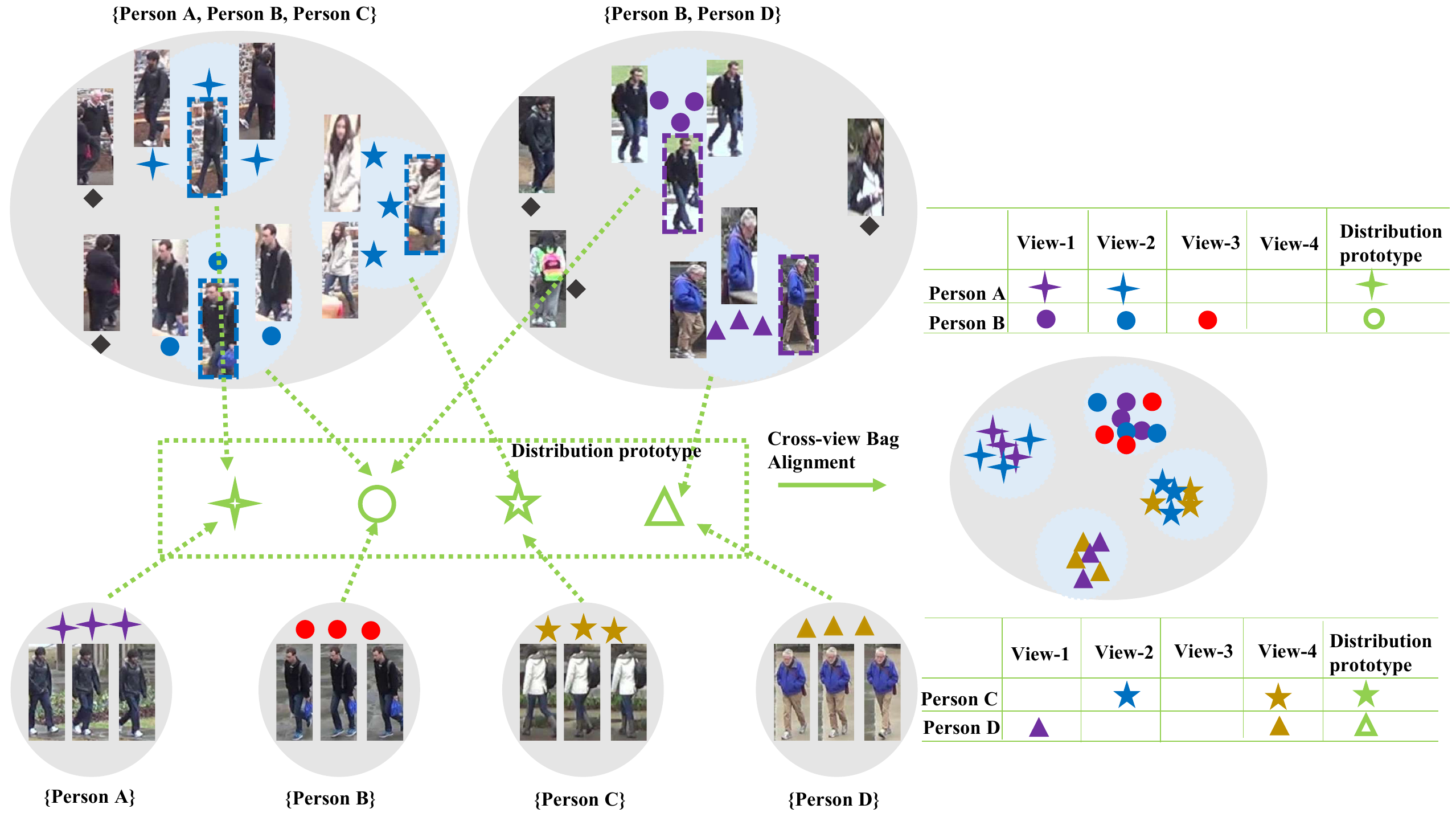}}
  \caption{Illustration of cross-view bag alignment. \jingkeee{The potential matched instances of the same person between bags from all the camera views are denoted by the same shape. The different camera views are represented by different colors. By performing cross-view bag alignment, the label distributions of these instances belonging to the same person are aligned w.r.t. the corresponding distribution prototype such that their features can be compact between each other in the learned feature space.} }
  \label{fig:cross} 
\end{figure}

\subsection{Deep Cross-view MIML Model}
Summarizing the above analysis, we obtain the \textbf{\textit{Cross-view Multi-label Multi-Instance learning (CV-MIML)}} method described below:
\begin{equation}\label{eq:ca-miml}
\begin{aligned}
\mathcal{L}_{CV-MIML}= \mathcal{L}_{C} +  \delta(\mathcal{L}_{IA}  +  \mathcal{L}_{CA} +  \mathcal{L}_{E})\mjk{,}
\end{aligned}
\end{equation}
\jkk{where $\delta$ controls the weight \ws{and contribution of $\mathcal{L}_{IA}$, $\mathcal{L}_{CA}$} and $\mathcal{L}_{E}$ to the whole CV-MIML loss. }\jingkee{ By incorporating the intra-bag alignment and the cross-view bag alignment, the label distributions of intraclass instances are aligned not only within the same video (bag) but also between videos (bags) across camera views, so that the intra-class instances \mjk{can} be compact \mjk{between} each other in the learned feature space.}
Here, $\mathcal{L}_{E}$ is an \textit{entropy regularization} term.
In the learning process, we expect that each instance can be ideally partitioned into a certain class (i.e., the known classes or a novel class).
For a weakly labeled bag in the gallery set, there may exist a certain number of instances far away from all the data groups that are formed in the intra-bag and cross-view bag alignment process. We call these instances \textit{outlier instances}. This designation indicates that these outlier instances probably do not approach any of the known identity classes. To alleviate the effect of these outlier instances, we design an {entropy regularization} term as follows:
\begin{equation}\label{eq:entropy}
\begin{aligned}
\mathcal{L}_{E}= \frac{1}{N_E} \sum \limits_{X_{b} \in \mathcal{X}_g} \sum \limits_{i\in \{1,\cdots,n_b\}} \sum \limits_{c \in \{0,\cdots,C\}} (-\tilde{y}_{b,i}^c \log  \tilde{y}_{b,i}^c ).
\end{aligned}
\end{equation}
\jingke{Reducing the entropy in Eq.(\ref{eq:entropy}) is to facilitate the outlier instances to be affiliated to a certain class.}
We now embed the proposed CV-MIML method into a deep neural network to form an end-to-end framework of the \textit{Deep CV-MIML} \mjk{model} that can learn coherent features and a robust MIML classifier simultaneously.

\subsection{Implementation Details}
To \ws{implement} our proposed model, we adopt Resnet-50 \cite{he2016deep} as our basic CNN for feature extraction.
\jingk{The \ws{fully-connected} layer in Resnet-50 is replaced by our MIML classifier.}
All input images are resized to $256\times128$.
\jingke{
The values of hyperparameters $\gamma$, $K$ and $\alpha$ are set by cross validation on the validation set. The parameter $\delta$ in Eq.(\ref{eq:ca-miml}) is designed
}\wss{as a function of $t$ that varies with time. Specifically, we let $\delta= w(t)$; the value of $w(t)\subseteq [0,1]$ initially increases with time and then reaches saturation and remains at the maximum value \cite{laine2016temporal}, which helps enhance the reliability of the model used in deep neural networks.}
The bounding boxes we used were automatically generated by the Mask R-CNN algorithm \cite{he2017mask} in advance for \wsh{the genuine WL-DukeMTMC-REID dataset}.
Indeed, many false positive bounding boxes are detected. 
\wsh{To exclude these distractors,} each bounding box is assigned a confidence score that indicates the possibility of that bounding box belonging to any of known classes.
We set a threshold for excluding the samples with confidence scores below the threshold. The confidence score is obtained from a deep network that is pretrained on the probe set.


\begin{table}[]
\renewcommand{\arraystretch}{0.8}
\centering
{\scriptsize
\resizebox{\columnwidth}{!}{
\begin{tabular}{c|c|c|c|c}
\hline
Dataset      & \begin{tabular}[c]{@{}c@{}}\# Camera\\  views\end{tabular} & \begin{tabular}[c]{@{}c@{}}\# Identities\\ (training/testing)\end{tabular} & \begin{tabular}[c]{@{}c@{}}\# Training \\ BBoxes \\ (probe/gallery)\end{tabular} & \begin{tabular}[c]{@{}c@{}}\# Testing \\ BBoxes \\ (probe/gallery)\end{tabular} \\ \hline
WL-DukeMTMC-REID  & 8                                                          & 880/1695                                                                    & 60,267/923,879                                                                     & 116,128/904,066                                                                    \\ \hline
WL-PRID2011  & 2                                                          & 100/100                                                                    & 11,201/8,191                                                                     & 12,129/8,512                                                                    \\ \hline
WL-iLIDS-VID & 2                                                          & 150/150                                                                    & 9,731/11,278                                                                     & 12,129/8,512                                                                  \\ \hline
WL-MARS      & 6                                                          & 631/630                                                                    & 38,324/460,236                                                                    & 36,988/472,978                                                                  \\ \hline
\end{tabular}}}
 \caption{\wush{Detailed information of the \jingke{one genuine and }three new simulated datasets for the weakly supervised person re-id.}}
\label{table:setting}
\vspace{-0.2cm}
\end{table}

\subsection{Testing}
\final{In the testing phase, the probe set and gallery set are formed in the same manner as the training set.}
\wusheng{Accordingly, our goal is to find the raw videos where the target person appears in the weakly supervised setting.} Specifically, for a bag $X_p$ in the \mjk{testing} probe set, the feature of this bag $\mathbf{x}_p$ is the average pooling of features over all image frames in this sequence. Then, the distance between the bag $\mathbf{x}_p$ in the \mjk{testing} probe set and the bag $\mathbf{x}_q$ in the \mjk{testing} gallery set is
\begin{equation}\label{eq:test}
D(p,q) = \min \{ d(\mathbf{x}_p, \mathbf{x}_{q,1}), d(\mathbf{x}_p, \mathbf{x}_{q,2}), ... , d(\mathbf{x}_p, \mathbf{x}_{q,n_p})\}
\end{equation}
where $d$ is the Euclidean distance operator.

\section{Experiments}

\subsection{Datasets and Settings}
The experiments were carried out on one genuine dataset
\wush{WL-DukeMTMC-REID} and three simulated datasets
WL-PRID 2011, WL-iLIDS-VID and WL-MARS.
The probe set contained all the target individuals to \wush{search for} in the
gallery set, and every known identity had trimmed image
sequences in the probe set for all datasets. The remainder
of the videos formed the gallery set. The four datasets were
constructed as follows.

\noindent \textbf{WL-DukeMTMC-REID} For the genuine WL-DukeMTMC-REID dataset, a set of raw videos DukeMTMC \cite{ristani2016MTMC} is available}. DukeMTMC is a multi-camera dataset
recorded outdoors \wss{at} the Duke University campus with 8 synchronized cameras.  The WL-DukeMTMC-REID dataset
was constructed from the first 50-minute raw HD videos.
 We split the raw videos into halves; the training set and testing set both have 25-minute raw videos. There are 880 and 1,695 identities \wss{appearing} in at least two camera views in the training and testing sets. To form the gallery set \wsh{for the WL-DukeMTMC-REID dataset},
we first randomly split the raw video into short video clips, with each clip comprising
\wsh{between 20 and 120} raw frames. Afterwards, we applied Mask-RCNN \wush{\cite{he2017mask}} to these video clips to detect individuals. Note that even though we obtain the bounding boxes, the \wsh{specific} label of each individual is still \wsh{unknown} for the gallery set. The details of this dataset is shown in Table \wush{\ref{table:setting}}.

For the three simulated datasets WL-PRID 2011, WL-iLIDS-VID and WL-MARS, the raw videos of these datasets are unavailable, so we formed the simulated datasets as follows. First, we randomly selected one trimmed image sequence for every known identity to form the probe set, and the rest of videos were used to form the gallery set. Then, $3\sim8$ sequences were randomly selected to form a weakly labeled bag, where only bag-level labels were available, \wsh{and} the \wsh{specific} label of each individual was \wsh{unknown}. In this way, we converted three existing video-based person re-id datasets PRID 2011~\cite{hirzer11}, iLIDS-VID~\cite{wang2014person} and MARS~\cite{zheng2016mars} to WL-PRID 2011, WL-iLIDS-VID and WL-MARS, respectively, for weakly supervised person re-id. The details of these new datasets are shown in Table \wush{\ref{table:setting}}.

\subsection{Evaluation Protocol}
To evaluate the performance of our method, the widely used cumulative matching characteristics (CMC) curve and mean average precision (mAP) are used for quantitative measurement.


\subsection{Evaluation of the Deep CV-MIML Model}

In our modeling of Deep CV-MIML, we introduce 1) the intra-bag alignment term, 2) the cross-view bag alignment term, and 3) an entropy regularization to eliminate outlier instances.
\final{To evaluate the efficiency of the each component, we adopt the MIML classifier (Eq. (\ref{eq:class_loss})) as the baseline method and conduct "baseline with IA", "baseline with CA" and "baseline with entropy"
for comparison to prove the effectiveness of all proposed components separately. The results are reported in Table~\ref{table:components}.}

Comparing the CV-MIML method to the baseline MIML classifier in Table~\ref{table:components}, it is clear that
our CV-MIML method is very effective in handling the weakly supervised person re-id problem. By simultaneously minimizing the intra-bag alignment and cross-view bag alignment loss functions, the same identities from the same camera \jkkk{view} and nonoverlapping camera \jkkk{views} could be more coherent with each other. These results represent a notable improvement in the rank-1 matching accuracy, e.g., \meng{10.79\%, 5.00\%, 18.67\% and 13.41\%} improvements were observed on the WL-DukeMTMC-REID, WL-PRID 2011, WL-iLIDS-VID and WL-MARS datasets, respectively. Considering mAP, we also obtain \meng{8$\sim$\wss{14}\%} improvement on these four weakly labeled re-id datasets.

\begin{table}[!tp] 
\renewcommand{\arraystretch}{0.8}
	\centering	
    {\scriptsize
		\begin{tabular}
          {p{2.5cm}<{\centering}|*{4}{p{0.6cm}<{\centering}|}{p{0.6cm}<{\centering}}}
            \hline
            \tiny{\textbf{{WL-DukeMTMC-REID}}} & r=1 & r=5 & r=10 & r=20 & mAP\\
            \hline
            CV-MIML  & \textbf{78.05}  & \textbf{90.50}  & \textbf{93.75} & \textbf{95.99} &\textbf{59.53}\\
            baseline + IA & 74.69 & 88.50 & 92.15 & {94.81} &56.97\\
            baseline+CA & 72.92 & 87.96 & 92.04 & 94.75 &55.30\\
            baseline+entropy & 70.56 & 85.90 & 90.15 & 92.68 & 53.05\\
            baseline  & 67.26 & 84.90 & 89.50 & 92.68 &50.96\\
            \hline
            \hline

            \textbf{{WL-PRID2011}} & r=1 & r=5 & r=10 & r=20 & mAP\\
            \hline
           CV-MIML  & \textbf{72.00}  & \textbf{89.00}  & {95.00} & \textbf{99.00} &\textbf{70.78}\\
            baseline+IA & 69.00 & \textbf{89.00} & {93.00} & {98.00} &65.89\\
            baseline+CA & 68.00& 87.00 & \textbf{96.00} & {98.00}  &63.72\\
            baseline+entropy & 70.00& \textbf{89.00} &\textbf{96.00} & \textbf{99.00} &67.32\\
            baseline  & 67.00 &{86.00} &{95.00} & {97.00} &62.87\\
            \hline
            \hline

            \textbf{{WL-iLIDS-VID}} & r=1 & r=5 & r=10 & r=20 & mAP\\
            \hline
            CV-MIML  & \textbf{60.00}  & {80.00}  & {87.33} & \textbf{96.67} &\textbf{56.01}\\

            baseline+IA & 55.33 & \textbf{80.67} & \textbf{89.33} & 95.33 &53.78\\
            baseline+CA & 52.67 & {78.00} & {88.00} & 95.33 &50.58\\
            baseline+entropy & 44.67 & {69.33} & 81.33 & {92.67} & 44.99\\
            baseline  & 41.33 & 70.00 & 83.33 & 94.67 &42.26\\
            \hline
            \hline

            \textbf{{WL-MARS}} & r=1 & r=5 & r=10 & r=20 & mAP\\
            \hline
            CV-MIML  & \textbf{66.88}  & \textbf{82.02}  & \textbf{87.22} & \textbf{91.48} &\textbf{55.16}\\
            baseline+IA & 62.15 & 80.44 & 85.80 & 89.75 &50.27\\
            baseline+CA & 63.09 & {79.97} & {84.23} & {88.96} &50.61\\
            baseline+entropy & 60.88 & 79.34 & 85.49 & 89.43 & 49.13\\
            baseline  & 53.47 & 71.77 & 79.02 & 85.49 &40.31\\
            \hline

    \end{tabular}}
    \caption{\final{Ablation study of the proposed CV-MIML method.
     The matching accuracy values ($\%$) at rank(r) = 1, 5, 10, 20 and mAP are shown on the four datasets. The best results are shown in
black boldface font.}
	}\label{table:components}
\end{table}

Moreover, as reported in Table~\ref{table:components}, the ablation study indicates that \final{adopting} \wss{the intra-bag alignment term} will lead to a \final{significant rise} of \wss{the model performance} because the {intra-bag alignment} term facilitates forming a coherent clustered structure for instances of the same identity.
Additionally, \final{including} the \jkkk{cross-view bag alignment} term would also notably \final{increase} the performance of CV-MIML (with approximately \final{5\%, 1\%, 11\% and 10\%} \final{rise} of rank-1 matching accuracy on the WL-DukeMTMC-REID, WL-PRID 2011, WL-iLIDS-VID and
WL-MARS datasets, respectively) because the cross-view bag alignment is useful for making the features of the same identities from nonoverlapping camera views aligned to each other in the feature space.


Finally, Table~\ref{table:components} \wss{indicates} that the entropy regularization term also plays a significant role in our CV-MIML model, as \final{with} it, the effect of outlier instances \final{can} be eliminated, thus \final{boosting} the performance of our model.



\subsection{Comparison with State-of-the-Art MIML Methods}

In Table~\ref{table:arts}\wsh{,}
we report the comparison of our method to existing state-of-the-art MIML learning methods DeepMIML \cite{feng2017deep} and
MIMLfast \cite{huang2014fast}.
The DeepMIML \cite{feng2017deep} method is an end-to-end deep neural network that integrates the instance representation learning process into the MIML learning.
For a fair comparison, we reimplemented this method using the same CNN structure and the same training process. The
MIMLfast \cite{huang2014fast} approach is a conventional two-stage framework that first requires extracting the image
features and then learns a discriminative representation.
\jkkk{
In this study, we extracted the features from a Resnet-50 CNN that was pretrained on the 3DPeS \cite{baltieri20113dpes}, CUHK01 \cite{li2013locally}, CUHK03 \cite{li2014deepreid}, Shinpuhkan \cite{kawanishi2014shinpuhkan2014} and VIPeR \cite{gray2007evaluating} person re-id datasets}
and then performed the MIML learning based on the MIMLfast method.

\begin{table}[t] 
\renewcommand{\arraystretch}{0.8}
	\centering
    {\scriptsize
		\begin{tabular}
          {p{2.5cm}<{\centering}|*{4}{p{0.6cm}<{\centering}|}{p{0.6cm}<{\centering}}}
            \hline
            \tiny{\textbf{WL-{DukeMTMC-REID}}} & r=1 & r=5 & r=10 & r=20 & mAP\\
            \hline
            MIMLfast\cite{huang2014fast}  &13.63  &44.66  &55.69  &64.78  &10.05  \\
            DeepMIML\cite{feng2017deep} &{65.37} &{82.30}  &{86.90}  &{90.68} &{48.02}\\
            Deep CV-MIML & \textbf{78.05}  & \textbf{90.50}  & \textbf{93.75} & \textbf{95.99} &\textbf{59.53}\\

            \hline
            \hline

            \textbf{{WL-PRID2011}} & r=1 & r=5 & r=10 & r=20 & mAP\\
            \hline
            MIMLfast\cite{huang2014fast}  &29.00  &56.00  &72.00  &87.00  &31.66  \\
            DeepMIML\cite{feng2017deep} &{67.00} &\textbf{90.00}  &{94.00}  &{99.00} &{61.80}\\
            Deep CV-MIML & \textbf{72.00}  & {89.00}  & \textbf{95.00} & \textbf{99.00} &\textbf{70.78}\\

            \hline
            \hline

            \textbf{WL-{iLIDS-VID}} & r=1 & r=5 & r=10 & r=20 & mAP\\
            \hline
            MIMLfast\cite{huang2014fast}  &28.00  &58.67  &69.33  &78.67  &27.42  \\
            DeepMIML\cite{feng2017deep} &{44.00} &{70.00}  &{81.33}  &{89.33} &{43.49}\\
            Deep CV-MIML & \textbf{60.00}  & \textbf{80.00}  & \textbf{87.33} & \textbf{96.67} &\textbf{56.01}\\

            \hline
            \hline

            \textbf{WL-{MARS}} & r=1 & r=5 & r=10 & r=20 & mAP\\
            \hline
            MIMLfast\cite{huang2014fast}  &20.50  &37.22  &43.06  &52.05  &11.39  \\
            DeepMIML\cite{feng2017deep} &{47.16} &{70.19}  &{76.18}  &{81.07} &{36.59}\\
            Deep CV-MIML  & \textbf{66.88}  & \textbf{82.02}  & \textbf{87.22} & \textbf{91.48} &\textbf{55.16}\\

            \hline

    \end{tabular}}
    \caption{
		\ws{Comparison with state-of-the-art MIML methods. The best results are in black boldface font.
        }\label{table:arts}
	}
\end{table}

The comparison shows that the proposed Deep CV-MIML model outperformed the existing MIML methods.
The proposed Deep CV-MIML \mjk{model} clearly outperformed the second-best method DeepMIML on the four datasets.
Specifically, the extra gain of the rank-1 matching accuracy between the Deep CV-MIML network and the DeepMIML method is \meng{12.68\%, 5.00\%, 16.00\% and 19.72\%} on the WL-DukeMTMC-REID, WL-PRID 2011, WL-iLIDS-VID and
WL-MARS datasets, respectively.
\wss{
Moreover, comparing the proposed method to the Deep MIML method, the mAP matching gain on all datasets  can reach \meng{11.51\%, 8.98\%, 12.52\% and 18.57\%}  on the WL-DukeMTMC-REID, WL-PRID 2011, WL-iLIDS-VID and
WL-MARS datasets, respectively.}
These results indicate the advantage of our Deep CV-MIML model in handling the weakly supervised person re-id problem.
The better performance is mainly due to the newly designed intra-bag alignment term and cross-view bag alignment term. With these terms, the features of the same individual obtained from the same camera view and across nonoverlapping camera views can be more coherent, while the functions of these two terms are not considered in MIMLfast and DeepMIML.

\subsection{Comparison with Related Re-id Methods}

\jkk{
As existing supervised person re-id methods could not be applied to our weakly supervised setting directly, we compare our method to unsupervised person re-id methods, \wsh{such as} CAMEL \cite{yu2017cross}, PUL \cite{fan2018unsupervised} and the one-shot person re-id method called EUG \cite{wu2018exploit}.
Among the listed methods, the CAMEL method is a conventional two-stage framework that first requires extracting the image features and then learns an asymmetric representation.
PUL and EUG are progressive methods that alternate between \meng{assigning the pseudo-labels to the tracklets and training the CNN model according to these pseudo-labeled data samples}.}
\final{To further demonstrate the effectiveness of our method, we also compared with a state-of-the-art fully supervised approach PCB\cite{sun2018beyond}.}
The results are reported in Table \ref{table:arts-reid}. Compared to unsupervised or one-shot methods, the performance of these methods is consistently unsatisfactory in comparison to that of the proposed Deep CV-MIML model. \final{The table can also tell us that  the performance of our model (Deep CV-MIML) is comparable to the fully supervised model PCB on the WL-DukeMTMC-REID and WL-MARS datasets.}

\begin{table}[t] 
\renewcommand{\arraystretch}{0.8}
	\centering
    {\scriptsize
		\begin{tabular}
          {p{2.5cm}<{\centering}|*{4}{p{0.6cm}<{\centering}|}{p{0.6cm}<{\centering}}}
            \hline
            \tiny{\textbf{WL-{DukeMTMC-REID}}} & r=1 & r=5 & r=10 & r=20 & mAP\\
            \hline
            CAMEL \cite{yu2017cross}  &0.53  &0.77  &1.18  &3.24  &0.90\\
            PUL\cite{fan2018unsupervised}   &-  &-  &-  &-  &- \\
            EUG\cite{wu2018exploit} &{35.93} &{50.74}  &{59.41}  &{66.96} &{21.94}\\
            \hline
            Deep CV-MIML  & \textcolor{blue}{78.05}  & \textcolor{red}{90.50}  & \textcolor{red}{93.75} &\textcolor{blue}{95.99} &\textcolor{blue}{59.53}\\
            \hline
            PCB\cite{sun2018beyond}   &\textcolor{red}{79.82}  &\textcolor{blue}{90.38}  &\textcolor{blue}93.45  &\textcolor{red}{96.17}  &\textcolor{red}{62.09} \\

            \hline
            \hline

            \textbf{{WL-PRID2011}} & r=1 & r=5 & r=10 & r=20 & mAP\\
            \hline
            CAMEL \cite{yu2017cross}  &2.00  &11.00  &20.00  &44.00  &4.59  \\
            PUL\cite{fan2018unsupervised}    &32.00  &58.00  &71.00  &85.00  &35.28  \\
            EUG\cite{wu2018exploit} &{55.00} &{83.00}  &{93.00}  &{97.00} &{53.26}\\
             \hline
            Deep CV-MIML  & \textcolor{blue}{72.00}  & \textcolor{blue}{89.00}  & \textcolor{blue}{95.00} & \textcolor{red}{99.00} &\textcolor{blue}{70.78}\\
            \hline
            PCB\cite{sun2018beyond}   &\textcolor{red}{88.00}  &\textcolor{red}{97.00}  &\textcolor{red}{99.00}  &\textcolor{red}{99.00}  &\textcolor{red}{87.35} \\
            \hline
            \hline

            \textbf{WL-{iLIDS-VID}} & r=1 & r=5 & r=10 & r=20 & mAP\\
            \hline
            CAMEL \cite{yu2017cross}  &4.67  &16.00  &26.67  &43.33  &6.26  \\
            PUL\cite{fan2018unsupervised}   &20.00  &44.00  &59.33  &76.00  &22.56  \\
            EUG\cite{wu2018exploit} &{26.67} &{60.67}  &{72.00}  &{86.67} &{29.86}\\
 \hline
            Deep CV-MIML  & \textcolor{blue}{60.00}  & \textcolor{blue}{80.00}  &\textcolor{blue}{87.33} & \textcolor{red}{96.67} &\textcolor{blue}{56.01}\\
            \hline
            PCB\cite{sun2018beyond}   &\textcolor{red}{72.00}  &\textcolor{red}{89.33}  &\textcolor{red}{92.67}  &\textcolor{blue}96.00  &\textcolor{red}{69.87} \\

            \hline
            \hline

            \textbf{WL-{MARS}} & r=1 & r=5 & r=10 & r=20 & mAP\\
            \hline

            CAMEL \cite{yu2017cross}  &0.32  &1.10  &2.52  &5.52  &0.56  \\
            PUL\cite{fan2018unsupervised}   &-  &-  &-  &-  &-  \\
            EUG\cite{wu2018exploit} &{25.87} &{39.59}  &{46.21}  &{55.21} &{15.63}\\
 \hline
            Deep CV-MIML   & \textcolor{blue}{66.88}  & \textcolor{blue}{82.02}  & \textcolor{red}{87.22} & \textcolor{red}{91.48} &\textcolor{red}{55.16}\\
            \hline
            PCB\cite{sun2018beyond}  &\textcolor{red}{68.14}  &\textcolor{red}{84.07}  &\textcolor{blue}{86.28}  &\textcolor{blue}{90.54}  &\textcolor{blue}{54.18} \\

            \hline

    \end{tabular}}
    \caption{
		\ws{Comparison with related re-id methods. \final{The $1^{st}/2^{nd}$ best results are indicated in \textcolor{red}{red}/\textcolor{blue}{blue}.}}\label{table:arts-reid}
	}
\end{table}

\subsection{Hyperparameter Analysis}
\begin{figure}[t]
\subfigure[\ws{Parameter} $\delta$]{
    \label{fig:delta} 
   \includegraphics[width=0.48\columnwidth]{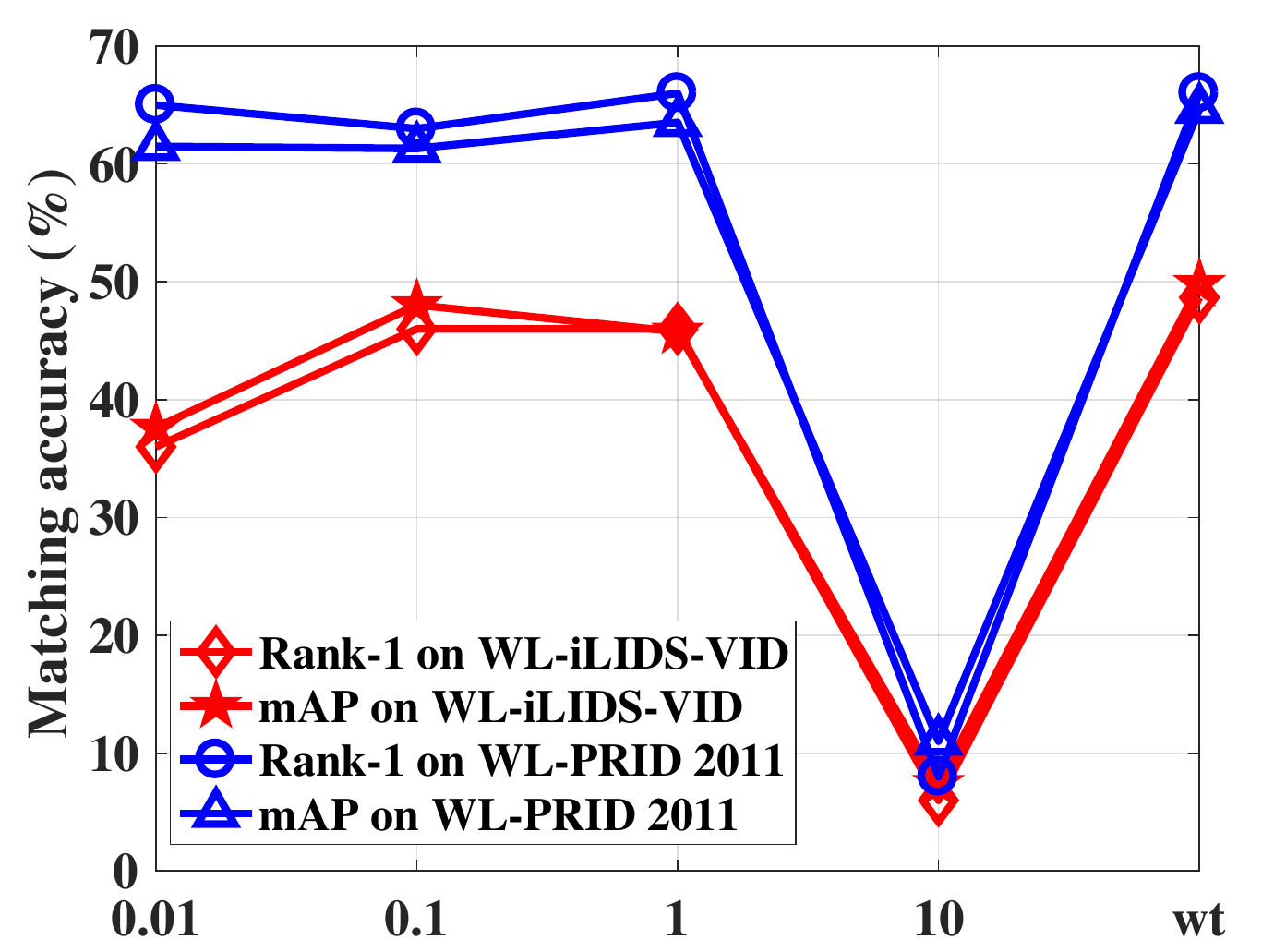}}
\subfigure[Parameter $K$]{
    \label{fig:k} 
    \includegraphics[width=0.48\columnwidth]{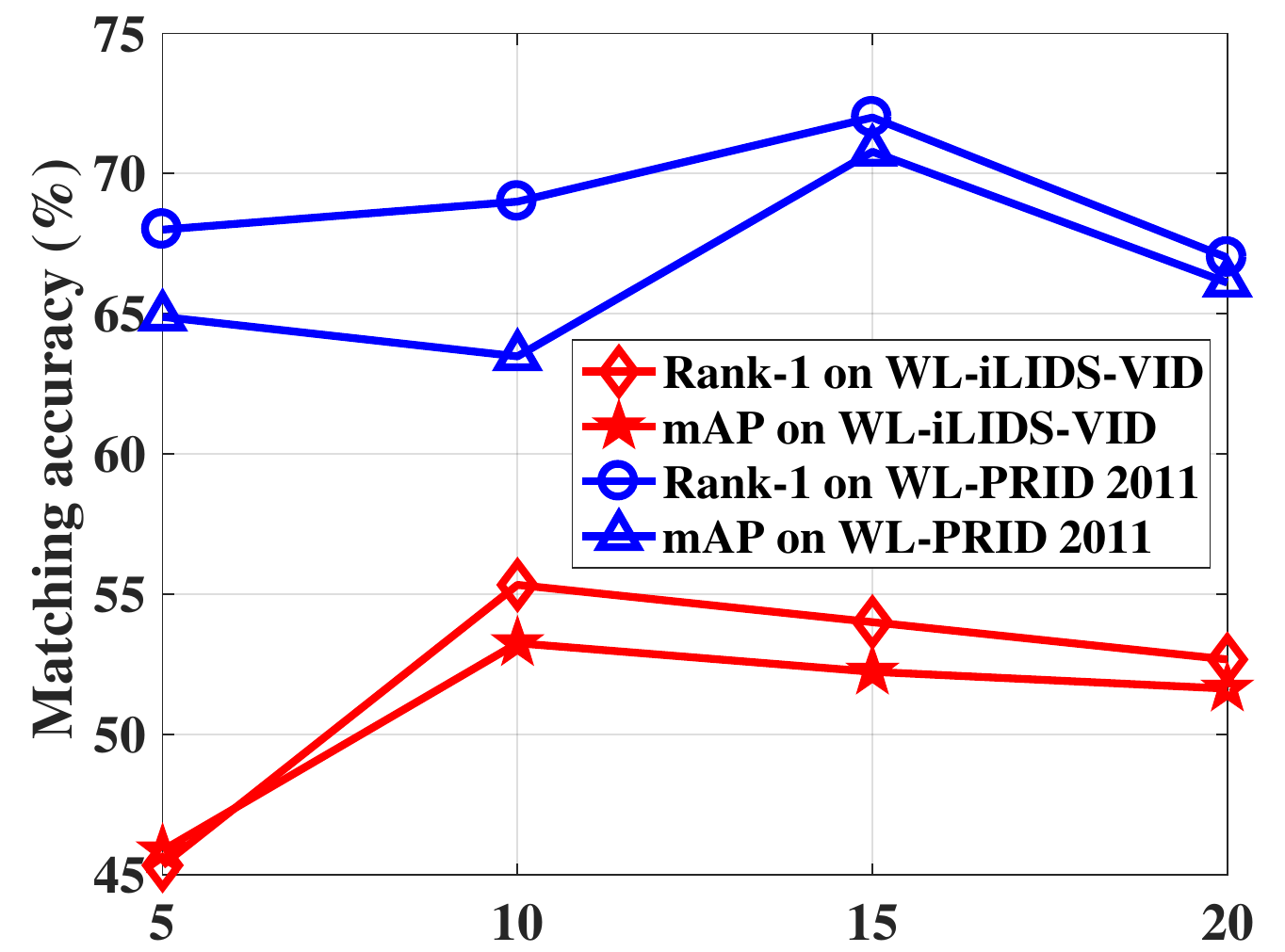}
   }
\subfigure[Parameter $\gamma$]{
    \label{fig:gamma} 
   \includegraphics[width=0.48\columnwidth]{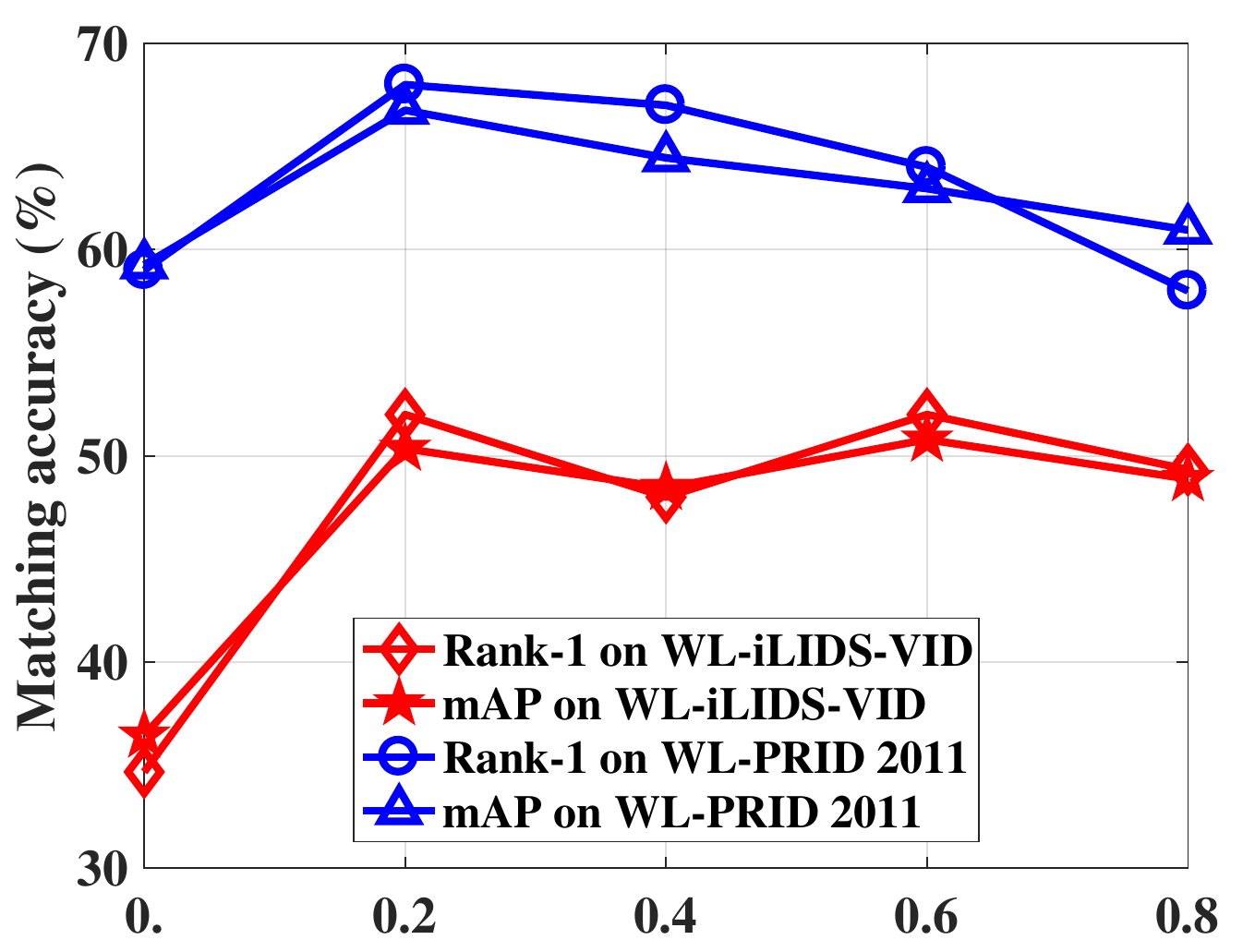}}
\subfigure[Parameter $\alpha$]{
    \label{fig:alpha} 
   \includegraphics[width=0.48\columnwidth]{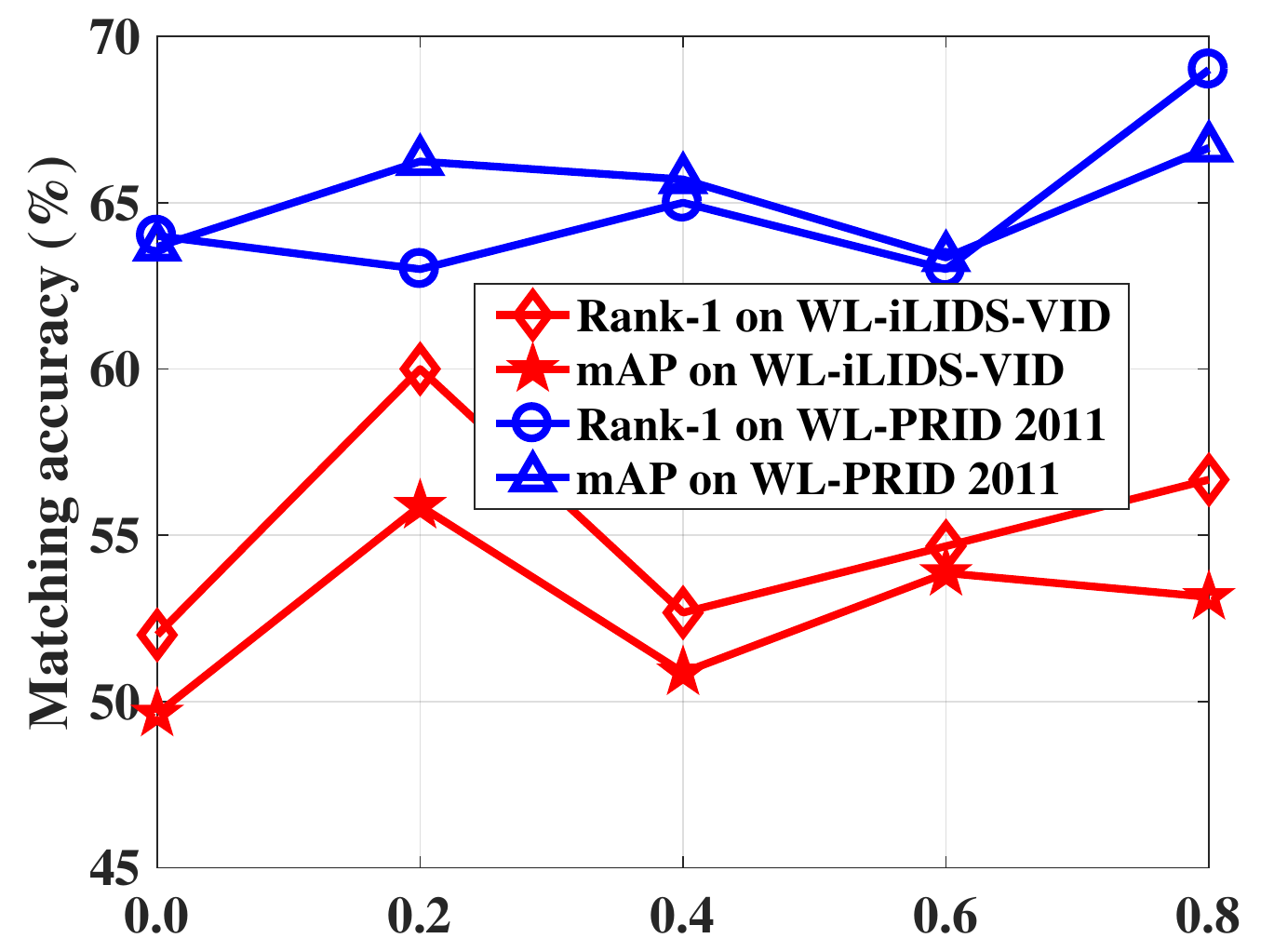}}
\caption{\ws{Performance illustrations for the Deep CV-MIML model with} different hyperparameters.}\label{fig:hyper-parameter}
\end{figure}

There are four hyperparameters involved in our CV-MIML formulation.
\jkk{\ws{
The trade-off parameter $\delta$ is used to
balance the weight of $\mathcal{L}_{IA}$, $\mathcal{L}_{CA}$ and $\mathcal{L}_{E}$ with respect to the overall CV-MIML loss \jkkk{in Eq. (\ref{eq:ca-miml})}.}
During training, we consider $\delta = w(t)$, a time-dependent function of time $t$.
To verify the advantage of this \wsh{approach}, we compared the performance to that of a fixed value of $\delta$, where $\delta= 0.01, 0.1, 1, 10$ to investigate the impact of $\delta$ on the overall performance
on the WL-PRID 2011 and WL-iLIDS-VID datasets. As shown in Figure \ref{fig:delta}, \meng{the time-dependent setting is preferable. The reason is that the reliability of the intra-bag alignment and cross-view bag alignment process is tightly related to the confidence of the re-id model \jingkee{by the seed instances selection and the distribution prototype calculation}. Additionally, the confidence of the re-id model is fairly low in the beginning and then steadily increases during the training procedure. 
Similarly, the weight parameter $\delta \in [0,1]$ initially increases during the early training stage, subsequently reaching saturation at approximately the maximum value 1 once the model has been sufficiently trained.
}


The group \ws{formed in the intra-bag alignment process is closely} related to parameters $K$ and $\gamma$. Parameter $K$ represents selecting the K-nearest neighbors 
in the feature space, and parameter $\gamma$ controls
\jingke{the number of} instances corresponding to those with the largest prior probabilities that should be shared with the same weak label.
The impact\ws{s} of $K$ and $\gamma$ are reported in Figure \ref{fig:k} and Figure \ref{fig:gamma}.
\meng{
The results suggest that the best performance can be reached on both datasets if $\gamma = 0.2$ and $K$ is approximately 15.
}

The impact of $\alpha$ is presented in Figure ~\ref{fig:alpha}. Parameter
$\alpha$ controls the impact of the historical distribution prototype when calculating the distribution prototype for the current epoch in Eq. \wsh{(\ref{eq:up_pct})}.
\meng{
The figure suggests that the performance with and without historical information in the calculation of the distribution prototype is distinct.
 Specifically,
the worst performance is observed if $\alpha=0$, i.e., involving the historical information that eliminates the bias of the current output is useful for the calculation of the distribution prototype.
}
}

\section{Conclusion}
We aim to remove the need for costly labeling efforts for conventional person re-id by considering weakly supervised person re-id modeling. In this weakly supervised setting,
no specific annotations of individuals inside gallery videos are necessary; the only requirement is the indication of whether or not a person appears in a given video. In such a setting, one can search for individuals and the videos that they appear in, given a (set of) probe person image(s).
We cast the weakly supervised person re-id problem as a multi-instance-multi-label (MIML) problem.
We develop a cross-view MIML (CV-MIML) method, which is able to mine potential intraclass variation in a bag and potential cross-view change between instances of the same person across bags from all camera views. Finally, CV-MIML is optimized by being embedded in a deep neural network. The experimental results have verified the feasibility of weakly supervised modeling for person re-id and have also shown the effectiveness of the proposed CV-MIML models.

\section*{Acknowledgment}
This work was supported partially by the National Key Research and Development Program of China (2018YFB1004903), NSFC(61522115), Guangdong Province Science and Technology Innovation Leading Talents (2016TX03X157), and the Royal Society Newton Advanced Fellowship (NA150459).

{\small
\bibliographystyle{ieee}
\bibliography{cites}
}

\end{document}